\documentclass[11pt]{article}
\usepackage{coling2020}
\usepackage{times}
\usepackage{url}
\usepackage{latexsym}
\usepackage{graphicx}
\usepackage[color=red!5]{todonotes}

\usepackage{tikz}
\usepackage{wrapfig}
\usepackage{xcolor}
\usepackage{soul}
\usepackage{paralist}

\newcommand{\hlc}[2][yellow]{{%
    \colorlet{foo}{#1}%
    \sethlcolor{foo}\hl{#2}}%
}

\definecolor{apricot}{rgb}{0.98, 0.81, 0.69}
\definecolor{cadetblue}{rgb}{0.37, 0.62, 0.63}
\definecolor{ceil}{rgb}{0.57, 0.63, 0.81}

\title{A Corpus for Argumentative Writing Support in German}
\author{Thiemo Wambsganss\textsuperscript{1}, Christina Niklaus\textsuperscript{1, 3}, Matthias Söllner\textsuperscript{2},\\ \textbf{Siegfried Handschuh\textsuperscript{1, 3}} \and \textbf{Jan Marco Leimeister\textsuperscript{1, 2}} \\
  \textsuperscript{1} University of St.Gallen\\
 \footnotesize{ {\tt \{thiemo.wambsganss, christina.niklaus, siegfried.handschuh, janmarco.leimeister\}{\tt @unisg.ch}}}\\
  \textsuperscript{2} University of Kassel\\
  \footnotesize{ {\tt \{soellner, leimeister\}{\tt@uni-kassel.de}}}\\
  \textsuperscript{3} University of Passau\\
  \footnotesize{ {\tt \{christina.niklaus, siegfried.handschuh\}{\tt@uni-passau.de}}}
\\}

\date{}

\begin{document}
\maketitle

\begin{abstract}
In this paper, we present a novel annotation approach to capture claims and premises of arguments and their relations in student-written persuasive peer reviews on business models in German language. We propose an annotation scheme based on annotation guidelines that allows to model claims and premises as well as support and attack relations for capturing the structure of argumentative discourse in student-written peer reviews. We conduct an annotation study with three annotators on 50 persuasive essays to evaluate our annotation scheme. The obtained inter-rater agreement of $\alpha=0.57$ for argument components and $\alpha=0.49$ for argumentative relations indicates that the proposed annotation scheme successfully guides annotators to moderate agreement. Finally, we present our freely available corpus of 1,000 persuasive student-written peer reviews on business models and our annotation guidelines to encourage future research on the design and development of argumentative writing support systems for students.
\end{abstract}
 
\section{Introduction}
\label{intro}
In today’s world most information is readily available. Consequently, the sole reproduction of information is losing attention. This results in a shift of job profiles towards interdisciplinary, ambiguous and creative tasks \cite{vomBrocke2018FutureSystems}. Therefore, educational institutions need to evolve in their curricula when it comes to the compositions of skills and knowledge conveyed. In particular, teaching higher order thinking skills to students, such as critical thinking, collaboration or problem-solving, has become more important \cite{Fadel2015Four-dimensionalSucceed}. This has already been recognized by the Organization for Economic Co-operation and Development (OECD), which included these skills as a major element of their Learning Framework 2030 \cite{OECD2018The2030}. One subclass represents the skill of arguing in a structured, reflective and well-formed way.  %\cite{Toulmin1984IntroductionReasoning}. 
Argumentation is not only an essential part of our daily communication and thinking, but it also contributes significantly to the competencies of communication, collaboration and problem-solving \cite{Kuhn1992ThinkingArgument}. Building on studies by Aristotle, the ability to form convincing arguments is recognized as the foundation for persuading an audience of novel ideas, and it plays a major role in strategic decision-making and analyzing different standpoints.
To develop skills such as argumentation, it is of great importance for the individual student to receive continuous feedback throughout their learning journey, also called formative feedback \cite{Black2009DevelopingAssessment,Hattie2007TheFeedback}. However, this is naturally hindered due to traditional large-scale lectures and due to the growing field of distance learning scenarios such as massive open online courses (MOOCs) \cite{Seaman2018HigherGroup}. In fact, educational institutions, such as universities, face the challenge of providing individual formative feedback effectively \cite{Fortes2010DealingChallenge}, since every student would need a personal tutor to have an optimal learning environment for learning how to argue \cite{vygotsky1980mind}. 

One possible path for providing individual feedback is to leverage recent developments in Computational Linguistics in the form of computer-assisted writing which enables the development of writing support systems that provide tailored feedback about textual documents \cite{song-etal-2014-applying,Stab2014AnnotatingEssays}. Argumentation Mining (AM), a research field in Computational Linguistics, aims at automatically identifying arguments in unstructured texts \cite{Lippi2015ArgumentationTrends}. %\textcolor{red}{Researchers use Argumentation Mining (AM) to develop algorithms that extract argumentative components and discourse structures from given texts} \cite{Lippi2015ArgumentationTrends}. 
%\textcolor{red}{Argumentation Mining models} mostly follow know argumentation theories \cite{Toulmin1984IntroductionReasoning,Walton2008ArgumentationSchemes}.
An argument is a set of statements made up of three parts: a claim, a set of evidence or premises (e.g., facts) and an inference from the evidence to the claim \cite{Toulmin1984IntroductionReasoning}. Claim and premise represent the argument components. The claim is the central component of an argument, representing an arguable text unit, while the premises are propositions that either support or attack the claim, underpinning its plausibility. Support and attack are argumentative relations that model the discourse structure of arguments. In the identification of these argumentation structures, two main tasks can be distinguished: 
\begin{itemize}
    \item \textbf{Argument component classification}: classification of argumentative text into claims and premises
    \item \textbf{Argument relation classification}: identification of support and attack relationships between pairs of argument components
\end{itemize}
%Identifying these argumentation structures usually involves three sub tasks \cite{Lawrence2019ArgumentSurvey}, such as identification of argumentative text \cite{Moens2007AutomaticTexts}, classification of argumentative components \cite{Stab2014AnnotatingEssays} and analysing argumentative discourse \cite{Stab2017ParsingEssays}. 
%However, the development of argumentative writing support systems for students based on Argumentation Mining is rather scarce in literature \cite{Stab2017ParsingEssays,Wambsganss2019TowardsTool}. One reason is that in order to design an accurate writing support system based on supervised machine learning approaches scientists need reliably annotated corpora of domain-related annotated texts from student-generated content to train a supervised machine learning model for extracting arguments \cite{Stab2014AnnotatingEssays,Stab2017ParsingEssays,Lawrence2019ArgumentSurvey}.

To design, train and test AM algorithms, the availability of high-quality labeled corpora is crucial. Therefore, numerous prior works have dealt with creating annotated data sets. However, they are all limited to a particular genre, ranging from well-structured legal \cite{Palau09,Ashley13} and scientific documents \cite{Kirschner15,Houngbo}, to rather ambiguous, vague and less formal social web content \cite{Wachsmuth14,Aharoni14,Cabrio14,Habernal15}. Corpora that are applicable for the design and development of argumentative writing support systems are scarce \cite{Stab2017ParsingEssays,Wambsganss2019TowardsTool,Wambsganss2020ALSkills}. The only collection from the education domain that is annotated for argumentative discourse structures was presented in \newcite{Stab2014AnnotatingEssays}. It is composed of 90 persuasive essays written by students in English language and later extended to include 402 essays \cite{Stab2017ParsingEssays}. However, these corpora are 1) annotated in English language only and 2) not derived from a specific learning scenario which would leverage the effective use for an argumentative writing support system. Consequently, there is a \textit{lack of linguistic corpora} for training models that provide students with adaptive feedback about the quality of their argumentation in common scenarios in large-scale lectures or the growing field of MOOCs \cite{Seaman2018HigherGroup}, such as peer reviews where students provide each other argumentative feedback on a specific task, e.g., on a previously developed business model \cite{Rietsche2019InsightsSkill}.
%This can be leveraged not only to provide students argumentative writing support in order to help them to write more persuasive peer reviews but also for a much wide ranger of writing support systems based on AM. To do so, a writing support system for students based on AM and a corpus of annotated peer reviews are necessary preconditions for the development \cite{Wambsganss2020ALSkills}. However, current literature falls short of a scientific annotation scheme for student peer reviews and a corpus which helps researchers and educators to drive research in the field of skill and argumentation learning for students in real-world learning scenarios through writing support systems.

Creating gold standards and test collections requires a formal representation model as well as corresponding annotation guidelines. In this paper, we introduce an argumentation annotation scheme for persuasive student-generated peer reviews extracted from a common learning scenario. Moreover, we present a corpus of 1,000 student-written peer reviews that are annotated for argumentation components and their relations. Our contribution is threefold: 1) we derive an annotation scheme for a new data domain for AM based on argumentation theory and previous work on annotation schemes for persuasive student essays \cite{Stab2017ParsingEssays,Stab2014AnnotatingEssays}, 2) we present an annotation study based on 50 persuasive peer reviews and three annotators to show that the annotation of student peer reviews is reliably possible, 3) we present our final and freely available corpus of 1,000 student peer reviews collected in our lecture about business innovation in German language. 
We therefore hope to encourage future research on student-generated argumentative texts and on writing support systems to train argumentation skills of students based on AM.

%\textcolor{red}{corpus will be made publicly available upon acceptance to support reproducibility blah blah}

%\textcolor{red}{The remainder of the paper is structured as follows: First, we provide an overview on related work on argumentation mining and existing annotated argumentation corpora. Next, in section 3,  we explain our annotation scheme based on argumentation theory. Section 4 presents the result of our annotation study and the final corpus creation. Finally, we discuss our results and how our corpora can be used to further develop adaptive argumentation tools in large-scale lecture scenarios.}  \todo{kann ggf. gestrichen werden}

\section{Related Work}
\label{relatedWork}

\subsection{Argumentation Mining}
\label{argumentationMining}
%An argument consists of one or more premises leading to exactly one conclusion. While argumentation connects together several arguments and thus, establishes chains of reasoning, claims are used as premises for deriving further claims. In that way, a regulated sequence of text with the goal of providing persuasive arguments for an intended conclusion or decision is constructed.
AM is a research field in Computational Linguistics, gaining momentum in a lot of areas, including the legal domain \cite{MochalesPalau09}, newswire articles \cite{Reed08,Deng15,Sardianos15}, user-generated web content \cite{Wachsmuth14,habernal-gurevych-2015-exploiting,Abbott16}, or online debates \cite{Cabrio14,EcklerKohler15}.
AM aims at automatically identifying arguments in unstructured textual documents based on the classification of argumentative and non-argumentative text units and the extraction of argument components and their relations. 
%Following \cite{Lippi2016ArgumentationTrends}, the most related subtasks of Argumentation Mining can be summed up as:
%\begin{itemize}
%\item \textbf{Argument identification}, which is concerned with identifying the argumentative parts in raw text and setting up its boundaries versus a non-argumentative text.
%\item \textbf{Argument component classification}, which is the subtask of which the primary purpose is to classify the components of the argument structure. Classifying an argumentative text into claims or premises is one popular way of tackling the target of this subtask.
%\item \textbf{Argumentative discourse analysis}, during this subtask, the researcher tries to identify the discourse relations between the various components existing in the argument. A typical example of this subtask is the identification of whether a support or an attack relationship exists between the claim and the premise.
%\end{itemize}
Recently, researchers have built increasing interest in intelligent writing assistance based on AM \cite{song-etal-2014-applying,Stab2014AnnotatingEssays,Stab2014IdentifyingEssays}, since it enables argumentative writing support systems that provide tailored feedback about arguments in student-generated texts. However, the effectiveness of using this technology in a certain learning scenario for educational purposes has rarely been assessed  \cite{stab-gurevych-2017-recognizing,Lippi2015ArgumentationTrends}, as argumentation corpora from student-generated texts in the field of education are rather uncommon \cite{Lawrence2019ArgumentSurvey}.

\subsection{Argument Annotated Corpora and Annotation Schemes}

As \newcite{Lawrence2019ArgumentSurvey} state, \textit{``one of the challenges faced by current approaches to argument mining is the lack of large quantities of appropriately annotated arguments to serve as training and test data.''}
Since the availability of labeled corpora is crucial for designing, training and evaluating AM algorithms, numerous prior works have dealt with creating annotated data sets, such as the Araucaria corpus \cite{Reed2008LanguageArgument}, the European Court of Human Rights (ECHR) corpus of \newcite{Mochales2008StudyLaw}, the Debatepedia corpus \cite{Cabrio2012NaturalApproach}, the ChangeMyView corpus \cite{egawa-etal-2019-annotating} or the persuasive essays corpus of \newcite{Stab2014AnnotatingEssays} with 90 essays and the corpus of \newcite{Stab2017ParsingEssays} with 402 persuasive student essays.
These corpora have been widely used for various AM tasks, such as the identification of argument components \cite{Rooney2012ApplyingMining}, corpus wide AM \cite{Ein-Dor2019CorpusSolution} or end-to-end AM \cite{Persing2016End-to-EndEssays}. 

Creating gold standards and test collections requires a formal representation model as well as corresponding annotation guidelines. While a number of well-defined models exist in the field of AM \cite{freeman2001argument,Perelman1971,pollock1995,Walton2015-WALACS-4,walton1996argumentation,Wambsganss2020UnlockingApproach}, there is no general argumentation annotation scheme across all domains and genres of texts. Instead, the proposed representations differ in granularity, expression power and categorization \cite{Lawrence2019ArgumentSurvey}. Therefore, conducting annotation studies with several annotators when introducing new annotation schemes is crucial for the quality of argumentation corpora. 

\subsection{Argument Annotated Corpora in Education}
With the exception of the corpora proposed in \newcite{Stab2014AnnotatingEssays} and \newcite{Stab2017ParsingEssays}, previously presented argument annotated datasets are not easily applicable for the development of argumentative writing support systems for students in a real-world case, since they are 1) not extracted from an educational learning scenario in which the annotation could be used for training a model that provides students feedback on the texts, and 2) often not annotated at the level of discourse \cite{Stab2017ParsingEssays,Lawrence2019ArgumentSurvey}, which is necessary for example to give students feedback on insufficiently supported claims. \newcite{Stab2014AnnotatingEssays} identified the lack of linguistic corpora in the domain of student-written texts for designing and developing argumentative writing support systems for adaptive feedback by leveraging AM techniques \cite{Stab2014AnnotatingEssays}. Therefore, they introduced an annotation scheme for annotating argument components and their relationships in persuasive student essays. Afterwards, several studies built on their corpus, including e.g., \newcite{Carlile2018GiveEssays} who take a subset of the essays and annotate their persuasiveness, or \newcite{Ke2018LearningEssays} who train a persuasiveness scoring model on them. However, the transfer of argumentation corpora to other educational domains (e.g., common learning scenarios such as peer reviews) and other languages falls short in current literature. 

\subsection{Essay Scoring Corpora}
Besides annotated datasets of argumentative student-written texts, several corpora exist in the field of automatic essay scoring. The goal of this task is to automatically rate textual documents in the form of holistic scores based on their content and form \cite{horbach2017fine}. Most corpora are built on student-written content, e.g. the Cambridge Learner Corpus (CLC) \cite{Yannakoudakis2011ATexts} with 1,244 English essays, the Swedish high school corpus with 1,702 essays \cite{Ostling2013AutomatedSwedish}, or the TOEFL11 corpus with 1,100 English essays written by language students \cite{Blanchard2013Toefl11:English}. However, the corpora are usually annotated with a holistic score on the level of full documents only, e.g., \textit{low, medium, high} in the TOEFL11 corpus, while specific argumentation structures are commonly ignored. In fact, the persuasiveness of essays is - if at all considered - usually only regarded as one sub variable of the overall document (e.g., in the annotations of \newcite{Ostling2013AutomatedSwedish} and \newcite{Yannakoudakis2011ATexts}). Consequently, these corpora lack applicability for the development of argumentative writing support systems. The argumentation quality is often only annotated as a qualitative score on a 1 to 3 range (e.g., \newcite{horbach2017fine}) and therefore not sufficient enough to train a sophisticated supervised machine learning model for argumentative writing assistants. Corpora from the field of automatic essay scoring usually neither distinguish between different types of argument components (e.g., claims or premises) nor are they built on a rich argumentation annotation scheme.
\newcite{Nguyen2018ArgumentEssays} demonstrated the value of AM for automated persuasive essay scoring by evaluating different AM features for improving essay scores. However, essay scoring corpora mostly do not focus on the annotation of argumentation relations, and therefore, disqualify for a foundation for sophisticated models for feedback on argumentative discourse through writing support systems (e.g., feedback on unsupported claims). 
We aim to address this literature gap by presenting and evaluating an annotation scheme as well as an annotated argumentation corpus built on student-written texts with the objective of enabling researchers to develop novel argumentative writing support systems for students.

\section{Corpus Construction}
\label{annotationScheme}
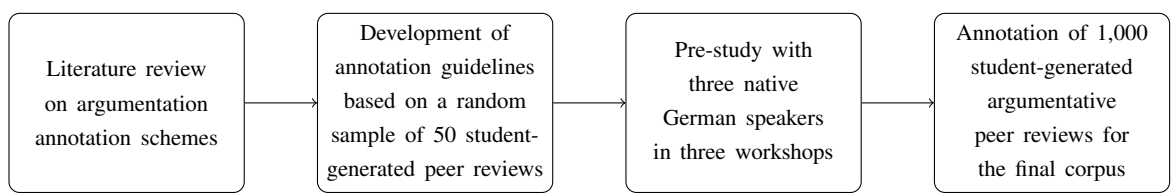
\begin{figure}[!ht]
\centering
\tikzstyle{myblock} = [rectangle, draw, minimum height=2.5cm, rounded corners] 
\begin{tikzpicture}[scale=0.95, transform shape]
    \node (1)[myblock, text width=3cm,align=center]{\small{Literature review on argumentation annotation schemes}};
    \node (2)[myblock,right =of 1,xshift=0cm, text width=3cm,align=center]{\small{Development of annotation guidelines based on a random sample of 50 student-generated peer reviews}};
    \node (3)[myblock,right =of 2,xshift=0cm, text width=3cm,align=center]{\small{Pre-study with three native German speakers in three workshops}};
    \node (4)[myblock,right =of 3,xshift=0cm, text width=3cm, align=center]{\small{Annotation of 1,000 student-generated argumentative peer reviews for the final corpus}};
    \draw[->] (1) -- node[below] {} (2);
    \draw[->] (2) -- node[below] {} (3);
    \draw[->] (3) -- node[below] {} (4);
    %\draw[->] (1) -- node[below] {arrow text} (2);
    % more arrows here
\end{tikzpicture}
\caption{Process of compiling a corpus of 1,000 annotated student-generated peer reviews.}
\label{fig:process}
\end{figure}

We compiled a corpus of 1,000 student-generated peer reviews in which students provide each other feedback on previously developed business models. Peer reviews are a modern learning scenario in large-scale lectures, enabling students to reflect on their content, receive individual feedback from peers and thus deepen their understanding of the content \cite{Rietsche2019InsightsSkill}. Moreover, they are easy to set up in traditional large-scale learning scenarios or in the growing field of distance-learning scenarios such as MOOCs \cite{Seaman2018HigherGroup}. This can be applied to train skills such as argumentation. However, since not many suitable corpora are available to provide argumentation feedback that A) contain annotated persuasive student peer reviews, B) consist of a sufficient size to be able to use trained models in a real-world scenario and C) follow an annotation guideline for guiding the annotators towards an adequate agreement, we propose an new annotation scheme to model argument components as well as argumentation relations that reflects the argumentative discourse structure in persuasive peer reviews. We based our annotation scheme on the model of \newcite{Toulmin1984IntroductionReasoning} and the studies of \newcite{Stab2014AnnotatingEssays,Stab2017ParsingEssays}. To build a reliable corpus, we followed a four step methodology, illustrated in Figure \ref{fig:process}: 1) We examined scientific literature and theory on how to model argumentation discourse structures in texts from different domains, 2) we randomly sampled 50 student-generated peer reviews and, on the basis of our findings from literature and theory, developed a set of annotation guidelines consisting of rules and limitations on how to annotate argumentation discourse structures, 3) we applied, evaluated and improved our guidelines with three native-speakers in three consecutive workshops to resolve annotation ambiguities, 4) we applied the final annotation scheme based on our 15 pages guidelines to a corpus of 1,000 student-generated peer reviews.\footnote{The annotation guidelines as well as the entire corpus can be accessed at \url{https://github.com/thiemowa/argumentative_student_peer_reviews}.}

\subsection{Data Source}
\label{Data}
We gathered a corpus of 1,000 student-generated peer reviews written in German. The data was collected in one of our mandatory business innovation lectures in a master program at our university. In this lecture, around 200 students develop and present a new business model for which they receive three peer reviews each. Here, a fellow student from the same course elaborates on the strengths and weaknesses of the business model and gives persuasive recommendations on what could be improved. We sampled a random subset of 1,000 of these reviews out of around 7,000 documents collected between 2014 and 2018.

\subsection{Annotation Scheme}
\label{annotation}
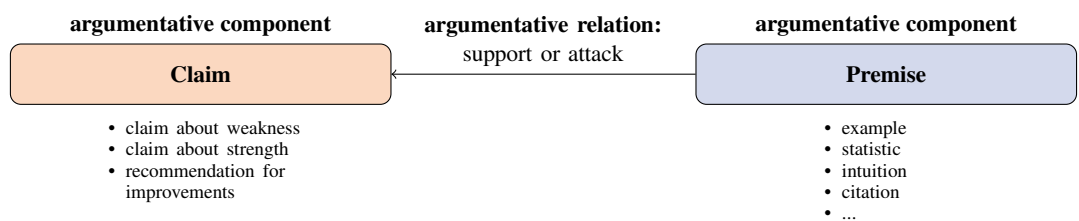
\begin{figure}[!htb]%{r}{11.5cm}
\centering
\tikzset{node distance = 0cm and 0cm}
\tikzstyle{myblock} = [rectangle, draw, minimum height=1cm, rounded corners] 
%\resizebox{0.68\textwidth}{5.2cm}{%
\begin{tikzpicture}[scale=0.8, transform shape]
    %\node (0)[text width=4cm,align=center, above =of 1]{\textbf{argumentative component}};
    \node (1)[myblock, text width=6cm,align=center, fill=apricot!80, label=above:\textbf{argumentative component}]{\textbf{Claim}};
    \node (2)[myblock,right =of 1,xshift=5cm, text width=6cm,align=center, fill=ceil!40, label=above:\textbf{argumentative component}]{\textbf{Premise}};
    %\node (3)[text width=4cm,align=center, above =of 2]{\textbf{argumentative component}};
    \node (4)[text width=4cm,align=center, below =of 1]{\small{\begin{compactitem}
        \item claim about weakness
        \item claim about strength
        \item recommendation for improvements
    \end{compactitem}}};
     \node (5)[text width=3cm,align=center, below =of 2]{\small{\begin{compactitem}
        \item example
        \item statistic
        \item intuition
        \item citation
        \item ...
    \end{compactitem}}};
    \draw[->] (2) -- node[above, text width=4cm, align=center] {\textbf{argumentative relation:} support or attack} (1);
    %\draw[->] (1) -- node[below] {arrow text} (2);
    % more arrows here
\end{tikzpicture}
%}
\caption{Argumentation annotation scheme including argument components (claim, premise) and argumentative relations (support or attack).}
  \label{fig:annotationscheme}
\end{figure}

Our objective is to model the argumentation discourse structures of student-generated peer reviews by annotating argumentation components and their relations. Most of the peer reviews in our corpus follow a similar structure. They describe several strengths or weaknesses of the business model under consideration, backing them up by examples, statistics, intuitions or citations. These strengths and weaknesses can also be formulated to make a certain recommendation for improvements of the business model. An argumentation component, e.g., a strength, weakness or suggestion is only regarded as a ``\textit{claim}'' if it contains a certain standpoint, which can also represent a complete sentence. Our basic annotation scheme is illustrated in Figure \ref{fig:annotationscheme}.

\subsubsection{Argument Components}
\label{Claims}
For argumentation components, we follow established models in argumentation theory which provide detailed definitions of argument components (e.g., \cite{Toulmin1984IntroductionReasoning,Walton2008ArgumentationSchemes,Weinberger2006ALearning,Perelman1971,pollock1995,walton1996argumentation,freeman2001argument,Walton2015-WALACS-4,van2016argumentation}. These theories generally agree that a basic argument consists of multiple components and that it includes a \textit{claim} that is supported or attacked by at least one \textit{premise}. Also in the domain of student-written peer reviews, we found that a \textit{claim} is the central component of an argument. It is a controversial statement (e.g., claiming a strength or weakness of a business model - see examples below) that is either true or false and should not be accepted by the receiver of the feedback without additional support or backing. The \textit{premise} supports the validity of the \textit{claim} (e.g., by providing a statistic, quote or a value-based intuition). It is a reason given by the author to persuade the reader of her \textit{claim}. Below are two examples of \textit{claims} from our corpus (\textit{1. }being a strength, \textit{2. }being a weakness) and their supporting \textit{premises}.\footnote{Since the original texts are written in German, we translated the examples to English for the sake of this paper.}

\begin{enumerate}
\item ``\underline{\hlc[apricot!80]{The value proposition is very well done.}}\textsubscript{claim (strength)} \textit{\hlc[ceil!40]{It is short and concise and the advantages or benefits are well highlighted.}}\textsubscript{premise (example)} \textit{\hlc[ceil!40]{As a customer I would like to try the product after reading it.}''}\textsubscript{premise (intuition)}
\item 
``\underline{\hlc[apricot!80]{Unfortunately, the value proposition canvas is very poorly filled in.}}\textsubscript{claim (weakness)} \textit{\hlc[ceil!40]{The points are far too little elaborated and far too little described or explained. The customer jobs should be described much more precisely.}''}\textsubscript{premise (example)}
\end{enumerate}

\subsubsection{Argumentative Relations}
\label{relations}
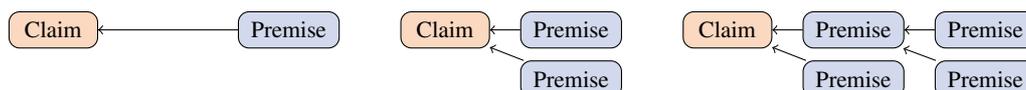
\begin{figure}[!ht]
\centering
\tikzset{node distance = 0.2cm and 0cm}
\tikzstyle{myblock} = [rectangle, draw, minimum height=0.6cm, rounded corners] 
\begin{tikzpicture}[scale=0.8, transform shape]
    \node (0)[myblock, text width=1.2cm,align=center, left =of 1, fill=apricot!80]{Claim};
    \node (1)[myblock, text width=1.4cm,align=center, fill=ceil!40]{Premise};
    %\node (a) [text width=1.5cm, align=center, below =of 0]{Example 1};
    \node (2)[myblock,right =of 1,xshift=1cm, text width=1.2cm,align=center, fill=apricot!80]{Claim};
    \node (3)[myblock, text width=1.4cm,align=center,xshift=0.5cm, right =of 2, fill=ceil!40]{Premise};
    \node (4)[myblock, text width=1.4cm,align=center, below =of 3, fill=ceil!40]{Premise};
    % \node (a) [text width=1.5cm, align=center, below =of 4]{Example 2};
    \node (5)[myblock,right =of 3,xshift=1cm, text width=1.2cm,align=center, fill=apricot!80]{Claim};
    \node (6)[myblock, text width=1.4cm,align=center,xshift=0.5cm, right =of 5, fill=ceil!40]{Premise};
    \node (7)[myblock, text width=1.4cm,align=center, below =of 6, fill=ceil!40]{Premise};
    \node (8)[myblock, text width=1.4cm,align=center,xshift=0.5cm, right =of 6, fill=ceil!40]{Premise};
    \node (9)[myblock, text width=1.4cm,align=center, below =of 8, fill=ceil!40]{Premise};
    % \node (a) [text width=1.5cm, align=center, below =of 7]{Example 3};
    
    \draw[->] (1) -- node[above, text width=0.5cm, align=center] {} (0);
    \draw[->] (3) -- node[above, text width=0.5cm, align=center] {} (2);
    \draw[->] (4) -- node[above, text width=0.5cm, align=center] {} (2);
    \draw[->] (6) -- node[above, text width=0.5cm, align=center] {} (5);
    \draw[->] (7) -- node[above, text width=0.5cm, align=center] {} (5);
    \draw[->] (8) -- node[above, text width=0.5cm, align=center] {} (6);
    \draw[->] (9) -- node[above, text width=0.5cm, align=center] {} (6);
    %\draw[->] (1) -- node[below] {arrow text} (2);
    % more arrows here
\end{tikzpicture}
\caption{Examples of possible argumentation relations. The arrow signifies a relation, which can either be \textit{support} or \textit{attack}. The rectangle denotes an argument component in the form of a \textit{claim} or \textit{premise}.}
  \label{fig:relations}
\end{figure}

The basic argumentation structure in our corpus of student-generated peer reviews consists of several claims, each independently supported by one or more premises. Since in our data domain weaknesses and strengths of a business model are discussed, the texts do generally not present a major claim as is the case for example in English essays annotated by \newcite{Stab2014AnnotatingEssays} and \newcite{Stab2017ParsingEssays}. Instead, the documents we deal with consist of a set of claims supported by one or more premises. However, premises not only \textit{support} a statement, but may also \textit{attack} a claim, e.g., when used as a stylistic device or to illustrate uncertainty in the argumentation. Hence, more complicated constellations of claims and premises are possible, in which a claim is supported by several different premises or by a chain of premises, in which each premise is in turn supported by another premise. In the same way, a claim can be supported by one premise and attacked by another, or supported by a premise which is attacked by another premise. Nevertheless, the simplest form consists of a claim supported by a single premise. To provide an overview, we illustrated three basic examples of annotated relations in our corpus (see Figure \ref{fig:relations}). Some statistics on the occurrences of different patterns in our dataset can be found in Table \ref{tab:patterns}.

\subsection{Annotation Process}
\label{annotationProcess}
\begin{figure}[!htb]
\centering
\tikzset{node distance = 0.5cm and 1cm}
\tikzstyle{myblock} = [rectangle, draw, minimum height=0.6cm, rounded corners] 
\begin{tikzpicture}[scale=0.9, transform shape]
    \node (0)[myblock, text width=6cm,align=center, left =of 1, label=left:\textbf{C1}, fill=apricot!80]{\footnotesize{\textbf{(1)} The value proposition is well done. Really a great idea.}};
    \node (1)[myblock, text width=8cm,align=center, xshift=4cm, yshift=-1cm, label=right:\textbf{P1}, fill=ceil!40]{\footnotesize{\textbf{(2)} Especially that it is not just a BrowserAddOn that shows where it is cheaper. But that you can also buy the products in a bundle via LivePrice without having to log on to all the different online stores and enter your data.}};
    \node (2)[myblock, text width=8cm,align=center, below =of 1, label=right:\textbf{P2}, fill=ceil!40]{\footnotesize{\textbf{(3)} The BMC describes that LivePrice packs and ships the products itself. [...]}};
    \node (3)[myblock, text width=6cm,align=center, below =of 0, yshift=-4.2cm, label=left:\textbf{C2}, fill=apricot!80]{\footnotesize{\textbf{(5)} In this case the customer would have to wait twice as long for his package as normally.}};
    \node (4)[myblock, text width=8cm,align=center, below =of 2, label=right:\textbf{P3}, fill=ceil!40]{\footnotesize{\textbf{(4)} This would mean, however, that LivePrice would first order the products from the provider who offers them at the lowest price and then pack them into a package.}};
    \node (5)[myblock, text width=6cm,align=center, below =of 3, label=left:\textbf{C3}, fill=apricot!80]{\footnotesize{\textbf{(6)} Wouldn't it perhaps make sense if the customers only ordered via LivePrice and the goods were then ultimately sent directly to the customer by the other store?}};
    \node (6)[myblock, text width=8cm,align=center, below =of 4, yshift=-2.2cm, label=right:\textbf{P4}, fill=ceil!40]{\footnotesize{\textbf{(7)} In this way LivePrice saves the very tedious logistics business and can then only charge a commission fee from the other stores.}};
    \node (7)[myblock, text width=8cm,align=center, below =of 6, label=right:\textbf{P5}, fill=ceil!40, yshift=-0.6cm]{\footnotesize{\textbf{(8)} This also keeps the costs within a manageable range.}};
    \node (8)[myblock, text width=6cm,align=center, below =of 5, yshift=-1.3cm, label=left:\textbf{C4}, fill=apricot!80]{\footnotesize{\textbf{(9)} In version 2 I would reconsider the logistics as described in 2.}};
    \node (9)[myblock, text width=6cm,align=center, below =of 8, label=left:\textbf{C5}, fill=apricot!80]{\footnotesize{\textbf{(10)} I would also go into detail about the advantages of a Premium Membership.}};
    \node (10)[myblock, text width=8cm,align=center, below =of 7, yshift=-1.4cm, fill=gray!20, label=right:\textbf{N}]{\footnotesize{\textbf{(11)} Only ad-free or even more advantages?}};
    \node (11)[myblock, text width=8cm,align=center, below =of 7, yshift=-2.4cm, label=right:\textbf{P6}, fill=ceil!40]{\footnotesize{\textbf{(12)} Since business models are more and more user-centered nowadays}};
    \node (12)[myblock, text width=6cm,align=center, below =of 9, label=left:\textbf{C6}, yshift=-1.4cm, fill=apricot!80]{\footnotesize{\textbf{(13)} I would think again about CRM.}};
    \node (13)[myblock, text width=8cm,align=center, below =of 11, label=right:\textbf{P7}, fill=ceil!40]{\footnotesize{\textbf{(14)} Unfortunately, chatbots are not yet ready to answer customer inquiries completely autonomously.}};

    \draw[->, very thick] (1) -- node[above,  align=center] {\footnotesize{support}} (0);
    \draw[->, very thick] (2) -- node[right,  align=center] {\footnotesize{support}} (4);
    \draw[->, very thick] (4) -- node[above,  align=center] {\footnotesize{support}} (3);
    \draw[->, very thick] (6) -- node[above,  align=center] {\footnotesize{support}} (5);
    \draw[->, very thick] (7) -- node[below,  align=center] {\footnotesize{support}} (5);
    \draw[->, very thick] (11) -- node[below,  align=center] {\footnotesize{support}} (12);
    \draw[->, very thick] (13) -- node[below left,  align=center] {\footnotesize{support}} (9);
    
    % more arrows here
\end{tikzpicture}
\caption{Fully annotated example of a peer review according to our annotation scheme and guidelines. The left column represents the claims (\textbf{C1 - C6}), while the premises are listed on the right (\textbf{P1 - P7}). \textbf{N} signifies a non-argumentative text unit.}
  \label{fig:annotation_example}
\end{figure}
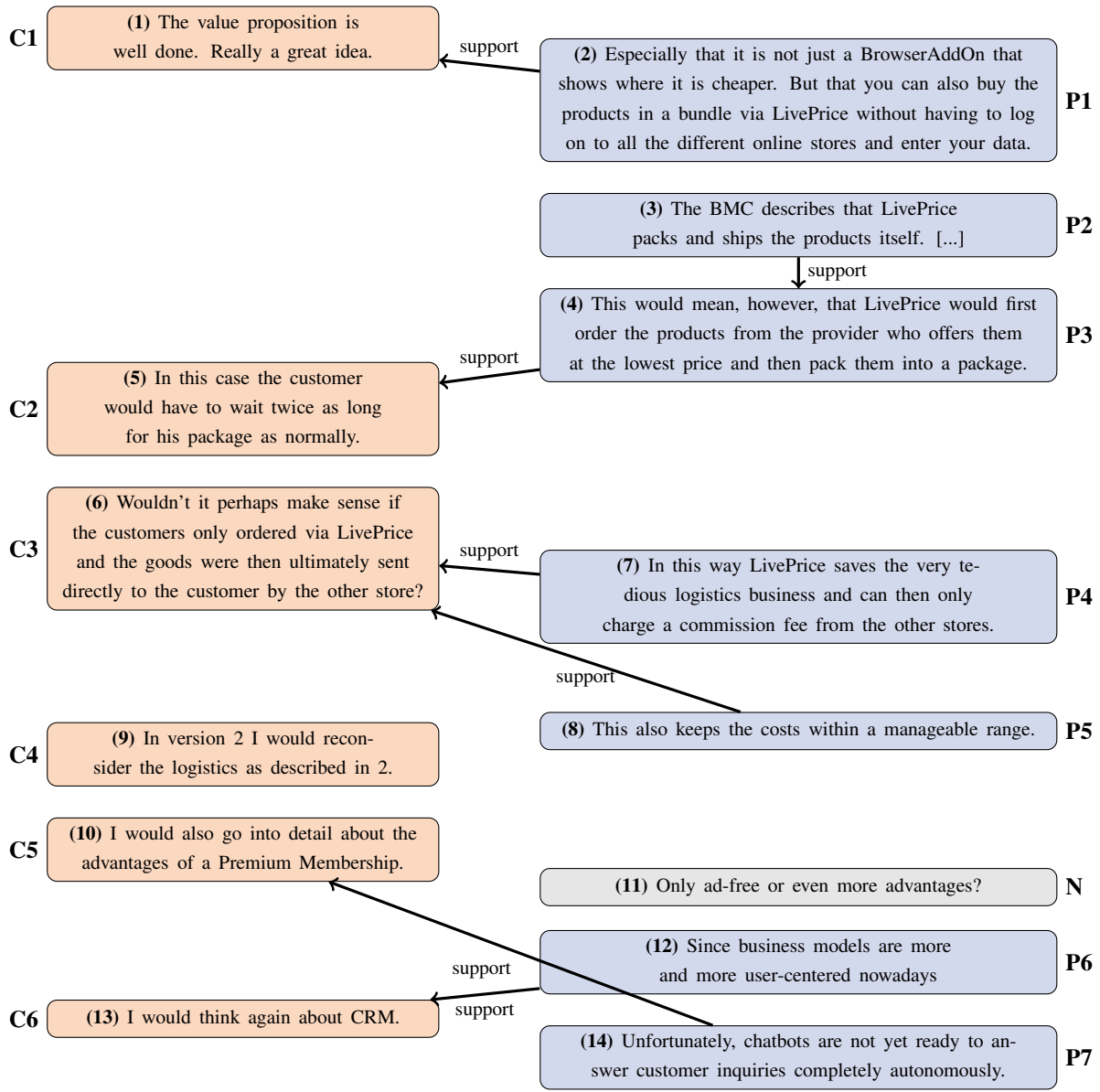

Three native German speakers annotated the peer reviews independently from each other for \textit{claims} and \textit{premises} as well as their \textit{argumentative relationships} in terms of support and attack, according to the annotation guidelines we specified. Inspired by \newcite{Stab2017ParsingEssays}, our guidelines consisted of 15 pages, including definitions and rules for what is an argument, which annotation scheme is to be used and how argument components and argumentative relations are to be judged. Several private training sessions and three team workshops were performed to resolve disagreements among the annotators and to reach a common understanding of the annotation guidelines. We used the brat rapid annotation tool, since it provides a graphical interface for marking up text units and linking their relations \cite{stenetorp-etal-2012-brat}. After the first 50 reviews were annotated by all three annotators, we calculated the inter-annotator agreement (IAA) scores. As we obtained satisfying results, we proceeded with a single annotator who marked up the remaining 950 documents. 
Following \newcite{Stab2014AnnotatingEssays}, we annotated argument components in terms of claims and premises. The specific component types, e.g., \textit{weakness, strength} for claims or \textit{intuition, statistic} for premise were not annotated in this study. However, we believe this would be a useful addition in future work.
In Figure \ref{fig:annotation_example} we display an example of an entire peer review with the corresponding annotations.

%\begin{figure}[!ht]
%  \centering
%  \includegraphics[width=0.85\textwidth]{coling2020/figures/relations.png}
%  \caption{Fully annotated example of a peer review according to our annotation scheme}
%  \label{fig:example}
%\end{figure}

\section{Corpus Analysis}
\label{corpusAnalysis}
We analyzed the results of the annotation process in order to examine (1) the \textit{reliability of the corpus} and
(2) the \textit{major disagreements} in argument component and relation annotations between the annotators. In addition, we calculated some statistics of the final corpus. %Our main findings are as follows:

\subsection{Inter-Annotator Agreement}
To evaluate the reliability of the argument component and argumentative relation annotations, we followed the approach of \newcite{Stab2014AnnotatingEssays}. 

\subsubsection{Argument Components}
With regard to the argument components, two strategies were used. Since there were no predefined markables, annotators not only had to identify the \textit{type of argument component}, but also its \textit{boundaries}. In order to assess the latter, we use Krippendorff's $\alpha\textsubscript{U}$ \cite{krippendorff2004}, which allows for assessing the reliability of an annotated corpus, considering the differences in the markable boundaries. To evaluate the annotators' agreement in terms of the selected category of an argument component for a given sentence, we calculate percentage agreement and two chance-corrected measures, multi-$\pi$ \cite{fleiss1971mns} and Krippendorff's $\alpha$ \cite{krippendorff80}. We decided to operate at sentence level, since only 20.56\% of the sentences in the corpus contain annotations of different argument components. Thus, evaluating the reliability at sentence level served as a good approximation of the IAA. At the level of individual sentences, 33.45\% contain a claim, 35.64\% a premise and 55.52\% a non-annotation. 20.56\% of the sentences contain several annotations, with a combination of a premise and a non-argumentative span (32.21\%) and a combination of a claim and a non-argumentative span (35.78\%) representing the majority of those cases. Only 4.08\% contain both a premise and a claim.
At the token level, the following class distribution is achieved: 43.54\% claim, 45.42\% premise and 11.04\% are not annotated. 

\begin{table}[!htb]%{r}{9.7cm}
\centering
\small
\begin{tabular}{|c|c|c|c|c|}
\hline
  & \% & Multi-$\pi$ & Krippendorff's $\alpha$ & Krippendorff's $\alpha\textsubscript{U}$\\ \hline\hline
 \textbf{Claim} & 0.7865 & 0.5547 & 0.5549 & 0.4404\\ \hline
 \textbf{Premise} & 0.7663 & 0.5106 & 0.5108 & 0.4776 \\ \hline
 \textbf{None} & 0.8572 & 0.6410 & 0.6412 & 0.3590\\ \hline
\end{tabular}
\caption{Inter-rater agreement of argument component annotations.}
\label{tab:agreement_components}
\end{table}

Table \ref{tab:agreement_components} displays the resulting IAA scores. We obtained an IAA of 78.65\% for the claims and 76.63\% for the premises. The corresponding multi-$\pi$ scores are 55.47\% and 51.06\%. Regarding Krippendorff's $\alpha$, a score of 55.49\% and 51.08\% is obtained, indicating a moderate agreement for both categories. With a score of 44.04\% and 47.76\%, the unitized $\alpha$ of both the claim and premise annotations is slightly smaller compared to the sentence-level agreement. Thus, the boundaries of argument components are less precisely identified in comparison to the classification into argument types. Yet the scores still suggest that there is a moderate level of agreement between the annotators. With a score of $\alpha\textsubscript{U}$=35.90\%, the boundaries of non-argumentative units are less reliably detected. In contrast, the agreement scores targeting the component types are considerably higher for the non-argumentative spans as compared to the claims and premises (85.72\%, 64.10\%, 64.12\%), indicating a substantial agreement between the annotators. Hence, we conclude that the annotation of the argument components in student-generated peer reviews is reliably possible.

\subsubsection{Argumentative Relations}

\begin{table}[!htb]%{r}{7.5cm}
\centering
\small
\begin{tabular}{|c|c|c|c|}
\hline
  & \% & Multi-$\pi$ & Krippendorff's $\alpha$ \\ \hline\hline
 \textbf{Support} & 0.9413 & 0.4903 & 0.4903 \\ \hline
 %\textbf{\textcolor{red}{Attack}} &  &  & \\ \hline
 \textbf{Non-support} & 0.9413 & 0.4903 & 0.4903 \\ \hline
\end{tabular}
\caption{Inter-rater agreement of argumentative relation annotations.}
\label{tab:agreement_relations}
\end{table}

To evaluate the reliability of the argumentative relations, we used the set of all pairs between argument components that were possible during the annotation task according to our annotation scheme, i.e. all pairs between a claim and a premise and between two premises. In total, the markables include 4,792 pairs of which 7.41\% are annotated as support relations and 0.35\% as attack relations. 92.24\% of the possible pairs were not identified by an annotator.
Since the number of attack relations is so small, we decided to focus on the support relations, distinguishing only between the two types \textit{support} and \textit{non-support}. We obtained an IAA of 94.13\% for both support and non-support relations. The corresponding multi-$\pi$ and Krippendorff's $\alpha$ scores both amount to 49.03\% (see Table \ref{tab:agreement_relations}).
Therefore, we conclude that argumentative relations, too, can be reliably annotated in student-generated argumentative peer reviews.

\subsection{Disagreement Analysis}

\subsubsection{Confusion Probability Matrices}
To analyze the disagreement between the three annotators, we created confusion probability matrices (CPM) \cite{cinkova-etal-2012-managing} for argument components and argumentative relations. A CPM contains the conditional probabilities that an annotator assigns a certain category (column) given that another annotator has chosen the category in the row for a specific item. In contrast to traditional confusion matrices, a CPM also allows for the evaluation of confusions if more than two annotators are involved in an annotation study \cite{Stab2014AnnotatingEssays}.

\begin{figure}[!ht]
\begin{minipage}{0.48\textwidth}
%\begin{table}[!ht]
\centering
\small
\begin{tabular}{|c||c|c|c|c|}
\hline
  & Claim & Premise & None \\ \hline\hline
 Claim & \textbf{0.5500} & 0.3296 & 0.1205 \\ \hline
 Premise & 0.3741 & \textbf{0.4570} & 0.1689 \\ \hline
 None & 0.1850 & 0.2285 & \textbf{0.5865} \\ \hline
\end{tabular}
\caption{CPM for argument component annotations.}
\label{tab:confusion_probability_matrix_component}
%\end{table}
\end{minipage}
\hfill
\begin{minipage}{0.48\textwidth}
%\begin{table}[!ht]
\centering
\small
\begin{tabular}{|c||c|c|c|}
\hline
  & Supports & None \\ \hline\hline
 Support & \textbf{0.3528} & 0.6472 \\ \hline
 Non-support & 0.0606 & \textbf{0.9394} \\ \hline
\end{tabular}
\caption{CPM for argumentative relation annotations.}
\label{tab:confusion_probability_matrix_relations}
%\end{table}
\end{minipage}
%\caption{Category ‘None’ indicates argument components that are not identified by an annotator.}
\end{figure}

While there is a broad agreement between the annotators in distinguishing  non-argumentative discourse units from argument components, the major disagreement is between claims and premises (Table \ref{tab:confusion_probability_matrix_component}). This result is in accordance with the findings in \newcite{Stab2014AnnotatingEssays}. This could be expected since a claim can also serve as premise for another claim, and it is difficult to distinguish these two concepts in the presence of reasoning chains. For instance, examples (1–3) establish a reasoning chain in which (1) is supported by (2) and (2) is supported by (3): %\todo{better example?}

\begin{compactitem}
    \item[(1)] \textit{``The client would have to wait twice as long as usual for his parcel.''}
    \item[(2)] \textit{``LivePrice orders the products from the provider that offers them at the best price and packs them.''}
    \item[(3)] \textit{``In the Business Model Canvas it is described that LivePrice packs the products itself and ships them.''}
\end{compactitem}

Considering the structure of this argumentation, (1) can be classified as a claim. However, if (1) is omitted, (2) becomes a claim that is supported by (3). Thus, the distinction between claims and premises depends not only on the context and the intention of the author but also on the structure of a specific argument \cite{Stab2014AnnotatingEssays}.

The CPM for argumentative relations (see Table \ref{tab:confusion_probability_matrix_relations}) reveals that there is a rather high confusion between support relations and none classified relations. This result is again in line with the findings of \newcite{Stab2014AnnotatingEssays}. In sum, according to our error analysis, the annotation of argumentative relations yields less reliable results than that of argument components.

\subsubsection{Manual Analysis}

We manually analyzed the most common differences in the annotated text spans between the annotators on a random sample of 10 documents that are composed of 292 sentences. Our findings are as follows:
while some annotators included \textit{clausal conjunctions}, such as ``because'', ``since'' or ``as'', in their annotations of premises, others did not mark them up. Moreover, in some cases \textit{phrases with verbs of reported speech or cognition}, such as ``I think that'' or ``I believe that'', were included in the annotations of argument components, whereas in other cases this phrasal type was ignored in the annotations. The same holds for \textit{introductory phrases} at the beginning of sentences or clauses, e.g. ``however'', ``moreover'', ``for example''. In addition, some annotators included \textit{punctuation marks} denoting the end of sentences or clauses in their annotated spans, while others did not. In addition, we encountered a larger number of cases where annotators \textit{mistakenly missed the first or last character(s) of a word} starting or ending, respectively, the annotated span. Often, the annotations were not even consistent within a single annotator.

%\todo{add percentage scores?}

%random sample of 10 documents (add number of sentences)
%most common differences between annotators:
%\begin{itemize}
%    \item clausal conjunctions (weil, da, daher, denn etc.)
%    \item punctuation at the end of a clause/sentence
%    \item introductory phrases, such as jedoch, zudem, z.B., d.h., auch, hier, teilweise, beispielsweise
%    \item satellite phrases such as ich denke, dass; ich nehme an, dass
%    \item mistakes where annotators mistakenly missed first/last (few) characters of a word starting/ending the annotated span
%\end{itemize}

%often not even consistent among 1 annotator...

\subsection{Corpus Statistics}

\begin{table}[!htb]%{r}{10.6cm}
\centering
\small
\begin{tabular}{|c||c|c|c|c|c|c|}
\hline
  & total & mean & standard deviation & min & max & median \\ \hline\hline
 \textbf{Sentences} & 20,217 & 20.217 & 9.4299 & 2 & 69 & 19.0  \\ \hline
 \textbf{Tokens} & 246,980 & 246.98 & 101.4358 & 40 & 720 & 236.0\\ \hline
\end{tabular}
\caption{Distribution of \textit{sentences} and \textit{tokens} in the created corpus. Mean, standard deviation, minimum, maximum and median refer to the number of sentences and tokens, respectively, per document.}
\label{tab:distribution_sentences_tokens}
\end{table}

Tables \ref{tab:distribution_sentences_tokens}, \ref{tab:distribution_components}, \ref{tab:distribution_relations} and \ref{tab:patterns} present some statistics of the final corpus. It consists of 1,000 student-written peer reviews in German that are composed of 20,125 sentences with 246,980 tokens in total. Hence, on average, each document has 20 sentences and 272 tokens. A total of 7,996 claims (31.64\%) and 8,479 premises (33.55\%) were annotated. 8,796 textual spans (34.81\%) were not identified as an argument component (``None''). On top of that, 8,291 support (97.03\%) and 254 attack (2.97\%) relationships were marked up by the annotators. The little percentage of attack relations are explainable due to the domain of our annotated texts. The nature of peer reviews is to provide feedback about a certain topic by highlighting strengths, weaknesses and suggesting improvements. Usually students used premises to support their claims, in fact, only about 3\% of the claims were further elaborated by discussing them more controversially through attacking premises.

\begin{table}[!ht]
\centering
\small
\begin{tabular}{|c||c|c|c|c|c|c|c|}
\hline
  & total & mean & standard deviation & min & max & median & percent\\ \hline\hline
 \textbf{Claim} & 7,996  & 7.996 & 3.6487 & 1 & 25 & 7.0 & 0.3164 \\ \hline
 \textbf{Premise} & 8,479 & 8.479 & 4.8346 & 0 & 26 & 8.0 & 0.3355\\ \hline
 \textbf{None} & 8,796 & 8.796 & 3.9545 & 2 & 26 & 8.0 & 0.3481 \\ \hline\hline
 \textbf{All} & 25,271 & 8.4237 & 4.1878 & 0 & 26 & 8.0 & 1.0 \\ \hline
\end{tabular}
\caption{Distribution of types of \textit{argument components} in the created corpus.} %Total describes the number of occurrences of the component in the document set. Mean refers to the average number of respective argument components per document. Standard deviation describes the corresponding amount of variation of the number of argumentative discourse units. Min denotes the minimum number of respective component found in a document, while max refers to the corresponding maximum number. Median signifies represents the value for which 50\% of observations a lower and 50\% are higher. Percent represents the percentage of the corresponding argumentative discourse units in the total set of documents.}
\label{tab:distribution_components}
\end{table}

\begin{table}[!ht]
\centering
\small
\begin{tabular}{|c||c|c|c|c|c|c|c|}
\hline
  & total & mean & standard deviation & min & max & median & percent\\ \hline\hline
 \textbf{Support} & 8,291 & 8.291 & 4.7771 & 0 & 26 & 7.0 & 0.9703 \\ \hline
 \textbf{Attack} & 254 & 0.254 & 0.6196 & 0 & 5 & 0.0 & 0.0297 \\ \hline\hline
% \textbf{non-support???} &  &  &  &  & & & \\ \hline\hline
 \textbf{All} & 8545 & 4.2725 & 5.2681 & 0 & 26 & 2.0 & 1.0 \\ \hline
\end{tabular}
\caption{Distribution of types of \textit{argumentative relations} in the created corpus.}
\label{tab:distribution_relations}
\end{table}

Table \ref{tab:patterns} presents the distribution of support relationships in our corpus. With 46.30\%, the majority of claims is not supported by any premise. 39.07\% of the claims are backed up by exactly one premise, while 10.93\% are supported by two premises. Patterns with more than two supporting premises rarely occur in our dataset. The majority of premises is not backed up by another premise (75.07\%). However, in 21.08\% of the cases, there is one supporting premise. In that way, more complex reasoning chains are established.

\begin{table}[!ht]
\centering
\small
\begin{tabular}{|c||c|c|c|c|c|c|}
\hline
  & \#0 support & \#1 support & \#2 support & \#3 support & \#4 support & $>$\#4 support \\ \hline\hline
 \textbf{Claim} & 0.4630 & 0.3907 & 0.1093 & 0.0230 & 0.0074 &  0.0066  \\ \hline
 \textbf{Premise} & 0.7507 & 0.2108 & 0.0289 & 0.0073 & 0.0019 & 0.0005 \\ \hline
% \textbf{non-support???} &  &  &  &  & & & \\ \hline\hline
\end{tabular}
\caption{Distribution of patterns of \textit{support relationships} in the created corpus. The columns denote the percentage of claims and premises, respectively, that are supported by the specified number of premises.} % (0, 1, 2, 3, 4, and more than 4).}
\label{tab:patterns}
\end{table}

\section{Application of the Corpus} %\todo[]{experiments?}
After constructing and analysing our corpus, we leveraged the novel data to train a machine learning model. Our objective was to embed a classification algorithm in the backend of a user-centered adaptive writing support system to provide students with formative argumentation feedback in the writing process. Therefore, we trained and tuned a model with different text features \cite{Stab2014AnnotatingEssays,Fromm2019}.
We performed a multiclass classification on the sentence level to detect the argument components and their relations. For argument component classification, we found that a Support Vector Machine (SVM) achieved the best results, with an accuracy of 65.4\% on the test set. Regarding the argumentative relation classification, a binary classification task, an SVM again achieved the best results on our corpus, obtaining an accuracy of 72.1\% on the test set.
A detailed description of the features and the embedding of our corpus in a user-centered adaptive writing support tool, as well as its effect on students' argumentation writing skills can be found in \newcite{Wambsganss2020ALSkills}.

\section{Conclusion}
\label{conclusion}
We introduce an argumentation annotation scheme as well as an annotated corpus of persuasive student-written peer reviews extracted from a real-world learning scenario which can be leveraged to provide students feedback on the quality of their argumentation (e.g., by highlighting insufficiently backed claims) through a writing support system. Regarding the educational domain, previously presented argument annotation schemes and argumentation corpora fall short on several aspects: they are not derived from a modern learning scenario, they do not follow a systematic methodology based on detailed inter-rater agreement studies or they do not include annotations of argumentative relations. To overcome these limitations, we present a corpus of 1,000 student-written peer reviews that are annotated for argument components and their relations. Our contribution is threefold: 1) we derive an annotation scheme for a new data domain for AM based on argumentation theory and previous work on annotation schemes for persuasive student essays \cite{Stab2014IdentifyingEssays,Stab2017ParsingEssays}, 2) we present an annotation study based on 50 persuasive peer reviews that were annotated by three native German speakers, demonstrating that the annotation of student-generated peer reviews is reliably possible, and 3) we present our final and freely available corpus of 1,000 student peer reviews collected in our lecture about business innovation in German language. We hope to encourage fellow researchers to leverage our annotation scheme and argumentation corpus to design and develop writing support systems for students in large-scale learning scenarios.

\bibliographystyle{acl}
\bibliography{coling2020}

\begin{thebibliography}{}

\bibitem[\protect\citename{Abbott \bgroup et al.\egroup }2016]{Abbott16}
Rob Abbott, Brian Ecker, Pranav Anand, and Marilyn Walker.
\newblock 2016.
\newblock Internet argument corpus 2.0: An sql schema for dialogic social media
  and the corpora to go with it.
\newblock In Nicoletta Calzolari~(Conference Chair), Khalid Choukri, Thierry
  Declerck, Sara Goggi, Marko Grobelnik, Bente Maegaard, Joseph Mariani, Helene
  Mazo, Asuncion Moreno, Jan Odijk, and Stelios Piperidis, editors, {\em
  Proceedings of the Tenth International Conference on Language Resources and
  Evaluation (LREC 2016)}, Paris, France. European Language Resources
  Association (ELRA).

\bibitem[\protect\citename{Aharoni \bgroup et al.\egroup }2014]{Aharoni14}
Ehud Aharoni, Anatoly Polnarov, Tamar Lavee, Daniel Hershcovich, Ran Levy, Ruty
  Rinott, Dan Gutfreund, and Noam Slonim.
\newblock 2014.
\newblock {A Benchmark Dataset for Automatic Detection of Claims and Evidence
  in the Context of Controversial Topics}.
\newblock In {\em Proceedings of the First Workshop on Argumentation Mining},
  pages 64--68, Baltimore, Maryland. Association for Computational Linguistics.

\bibitem[\protect\citename{Ashley and Walker}2013]{Ashley13}
Kevin~D. Ashley and Vern~R. Walker.
\newblock 2013.
\newblock Toward constructing evidence-based legal arguments using legal
  decision documents and machine learning.
\newblock In {\em Proceedings of the Fourteenth International Conference on
  Artificial Intelligence and Law}, ICAIL '13, pages 176--180, New York, NY,
  USA. ACM.

\bibitem[\protect\citename{Black and
  Wiliam}2009]{Black2009DevelopingAssessment}
Paul Black and Dylan Wiliam.
\newblock 2009.
\newblock {Developing the theory of formative assessment}.
\newblock {\em Educational Assessment, Evaluation and Accountability},
  21(1):5--31.

\bibitem[\protect\citename{Blanchard \bgroup et al.\egroup
  }2013]{Blanchard2013Toefl11:English}
Daniel Blanchard, Joel Tetreault, Derrick Higgins, Aoife Cahill, and Martin
  Chodorow.
\newblock 2013.
\newblock {Toefl11: a Corpus of Non-Native English}.
\newblock {\em ETS Research Report Series}, 2013(2):i--15.

\bibitem[\protect\citename{Cabrio and Villata}2012]{Cabrio2012NaturalApproach}
Elena Cabrio and Serena Villata.
\newblock 2012.
\newblock {Natural language arguments: A combined approach}.
\newblock {\em Frontiers in Artificial Intelligence and Applications},
  242:205--210.

\bibitem[\protect\citename{Cabrio and Villata}2014]{Cabrio14}
Elena Cabrio and Serena Villata.
\newblock 2014.
\newblock {Towards a Benchmark of Natural Language Arguments}.
\newblock {\em CoRR}, abs/1405.0.

\bibitem[\protect\citename{Carlile \bgroup et al.\egroup
  }2018]{Carlile2018GiveEssays}
Winston Carlile, Nishant Gurrapadi, Zixuan Ke, and Vincent Ng.
\newblock 2018.
\newblock {Give me more feedback: Annotating argument persuasiveness and
  related attributes in student essays}.
\newblock {\em ACL 2018 - 56th Annual Meeting of the Association for
  Computational Linguistics, Proceedings of the Conference (Long Papers)},
  1:621--631.

\bibitem[\protect\citename{Cinkov{\'a} \bgroup et al.\egroup
  }2012]{cinkova-etal-2012-managing}
Silvie Cinkov{\'a}, Martin Holub, and Vincent Kr{\'\i}{\v{z}}.
\newblock 2012.
\newblock Managing uncertainty in semantic tagging.
\newblock In {\em Proceedings of the 13th Conference of the {E}uropean Chapter
  of the Association for Computational Linguistics}, pages 840--850, Avignon,
  France, April. Association for Computational Linguistics.

\bibitem[\protect\citename{Deng and Wiebe}2015]{Deng15}
Lingjia Deng and Janyce Wiebe.
\newblock 2015.
\newblock {MPQA 3.0: An Entity/Event-Level Sentiment Corpus}.
\newblock In {\em Proceedings of the 2015 Conference of the North American
  Chapter of the Association for Computational Linguistics: Human Language
  Technologies}, pages 1323--1328, Denver, Colorado. Association for
  Computational Linguistics.

\bibitem[\protect\citename{Eckle-Kohler \bgroup et al.\egroup
  }2015]{EcklerKohler15}
Judith Eckle-Kohler, Roland Kluge, and Iryna Gurevych.
\newblock 2015.
\newblock On the role of discourse markers for discriminating claims and
  premises in argumentative discourse.
\newblock In {\em Proceedings of the 2015 Conference on Empirical Methods in
  Natural Language Processing}, pages 2236--2242, Lisbon, Portugal. Association
  for Computational Linguistics.

\bibitem[\protect\citename{Egawa \bgroup et al.\egroup
  }2019]{egawa-etal-2019-annotating}
Ryo Egawa, Gaku Morio, and Katsuhide Fujita.
\newblock 2019.
\newblock Annotating and analyzing semantic role of elementary units and
  relations in online persuasive arguments.
\newblock In {\em Proceedings of the 57th Annual Meeting of the Association for
  Computational Linguistics: Student Research Workshop}, pages 422--428,
  Florence, Italy, July. Association for Computational Linguistics.

\bibitem[\protect\citename{Ein-Dor \bgroup et al.\egroup
  }2019]{Ein-Dor2019CorpusSolution}
Liat Ein-Dor, Eyal Shnarch, Lena Dankin, Alon Halfon, Benjamin Sznajder, Ariel
  Gera, Carlos Alzate, Martin Gleize, Leshem Choshen, Yufang Hou, Yonatan Bilu,
  Ranit Aharonov, and Noam Slonim.
\newblock 2019.
\newblock {Corpus Wide Argument Mining -- a Working Solution}.
\newblock (1).

\bibitem[\protect\citename{Fadel \bgroup et al.\egroup
  }2015]{Fadel2015Four-dimensionalSucceed}
Charles Fadel, Bernie Trilling, and Maya Bialik.
\newblock 2015.
\newblock {\em {Four-Dimensional Education: The Competencies Learners Need to
  Succeed}}.

\bibitem[\protect\citename{Fleiss}1971]{fleiss1971mns}
J.L. Fleiss.
\newblock 1971.
\newblock {Measuring nominal scale agreement among many raters}.
\newblock {\em Psychological Bulletin}, 76(5):378--382.

\bibitem[\protect\citename{Fortes and
  Tchantchane}2010]{Fortes2010DealingChallenge}
Pauline~Carolyne Fortes and Abdellatif Tchantchane.
\newblock 2010.
\newblock {Dealing with large classes: A real challenge}.
\newblock {\em Procedia - Social and Behavioral Sciences}, 8:272--280.

\bibitem[\protect\citename{Freeman}2001]{freeman2001argument}
James~B Freeman.
\newblock 2001.
\newblock {Argument structure and disciplinary perspective}.
\newblock {\em Argumentation}, 15(4):397--423.

\bibitem[\protect\citename{Fromm \bgroup et al.\egroup }2019]{Fromm2019}
Hansjörg Fromm, Thiemo Wambsganss, and Matthias S{\"{o}}llner.
\newblock 2019.
\newblock {Towards a Taxonomy of Text Mining Features}.
\newblock In {\em European Conference of Information Systems (ECIS)}, pages
  1--12.

\bibitem[\protect\citename{Habernal and Gurevych}2015a]{Habernal15}
Ivan Habernal and Iryna Gurevych.
\newblock 2015a.
\newblock Exploiting debate portals for semi-supervised argumentation mining in
  user-generated web discourse.
\newblock In {\em Proceedings of the 2015 Conference on Empirical Methods in
  Natural Language Processing (EMNLP)}, pages 2127--2137, Lisbon, Portugal.
  Association for Computational Linguistics.

\bibitem[\protect\citename{Habernal and
  Gurevych}2015b]{habernal-gurevych-2015-exploiting}
Ivan Habernal and Iryna Gurevych.
\newblock 2015b.
\newblock Exploiting debate portals for semi-supervised argumentation mining in
  user-generated web discourse.
\newblock In {\em Proceedings of the 2015 Conference on Empirical Methods in
  Natural Language Processing}, pages 2127--2137, Lisbon, Portugal, September.
  Association for Computational Linguistics.

\bibitem[\protect\citename{Hattie and Timperley}2007]{Hattie2007TheFeedback}
John Hattie and Helen Timperley.
\newblock 2007.
\newblock {The Power of Feedback}.
\newblock {\em Review of Educational Research}, 77(1):81--112.

\bibitem[\protect\citename{Horbach \bgroup et al.\egroup
  }2017]{horbach2017fine}
Andrea Horbach, Dirk Scholten-Akoun, Yuning Ding, and Torsten Zesch.
\newblock 2017.
\newblock Fine-grained essay scoring of a complex writing task for native
  speakers.
\newblock In {\em Proceedings of the 12th Workshop on Innovative Use of NLP for
  Building Educational Applications}, pages 357--366.

\bibitem[\protect\citename{Houngbo and Mercer}2014]{Houngbo}
Hospice Houngbo and Robert Mercer.
\newblock 2014.
\newblock An automated method to build a corpus of rhetorically-classified
  sentences in biomedical texts.
\newblock In {\em Proceedings of the First Workshop on Argumentation Mining},
  pages 19--23, Baltimore, Maryland. Association for Computational Linguistics.

\bibitem[\protect\citename{Ke \bgroup et al.\egroup
  }2018]{Ke2018LearningEssays}
Zixuan Ke, Winston Carlile, Nishant Gurrapadi, and Vincent Ng.
\newblock 2018.
\newblock {Learning to give feedback: Modeling attributes affecting Argument
  persuasiveness in student essays}.
\newblock {\em IJCAI International Joint Conference on Artificial
  Intelligence}, 2018-July:4130--4136.

\bibitem[\protect\citename{Kirschner \bgroup et al.\egroup }2015]{Kirschner15}
Christian Kirschner, Judith Eckle-Kohler, and Iryna Gurevych.
\newblock 2015.
\newblock Linking the thoughts: Analysis of argumentation structures in
  scientific publications.
\newblock In {\em Proceedings of the 2nd Workshop on Argumentation Mining held
  in conjunction with the 2015 Conference of the North American Chapter of the
  Association for Computational Linguistics – Human Language Technologies
  (NAACL HLT 2015)}, pages 1--11.

\bibitem[\protect\citename{Krippendorff}1980]{krippendorff80}
Klaus Krippendorff.
\newblock 1980.
\newblock {\em Content Analysis: An Introduction to Methodology}.
\newblock Sage Publications, Inc., Beverly Hills, CA.

\bibitem[\protect\citename{Krippendorff}2004]{krippendorff2004}
Klaus Krippendorff.
\newblock 2004.
\newblock Measuring the reliability of qualitative text analysis data.
\newblock {\em Quality and Quantity}, 38(6):787--800, Dec.

\bibitem[\protect\citename{Kuhn}1992]{Kuhn1992ThinkingArgument}
Deanna Kuhn.
\newblock 1992.
\newblock {Thinking as Argument}.
\newblock {\em Harvard Educational Review}, 62(2):155--179, 7.

\bibitem[\protect\citename{Lawrence and Reed}2019]{Lawrence2019ArgumentSurvey}
John Lawrence and Chris Reed.
\newblock 2019.
\newblock {Argument mining: A survey}.
\newblock {\em Computational Linguistics}, 45(4):765--818.

\bibitem[\protect\citename{Lippi and
  Torroni}2015]{Lippi2015ArgumentationTrends}
Marco Lippi and Paolo Torroni.
\newblock 2015.
\newblock {Argumentation Mining: State of the Art and Emerging Trends}.
\newblock {\em IJCAI International Joint Conference on Artificial
  Intelligence}, 2015-Janua(2):4207--4211.

\bibitem[\protect\citename{Mochales and Moens}2008]{Mochales2008StudyLaw}
Raquel Mochales and Marie~Francine Moens.
\newblock 2008.
\newblock {Study on the structure of argumentation in case law}.
\newblock {\em Frontiers in Artificial Intelligence and Applications},
  189(1):11--20.

\bibitem[\protect\citename{Mochales~Palau and Ieven}2009]{MochalesPalau09}
Raquel Mochales~Palau and Aagje Ieven.
\newblock 2009.
\newblock Creating an argumentation corpus: do theories apply to real
  arguments? {A} case study on the legal argumentation of the {ECHR}.
\newblock In {\em Proceedings of the Twelfth International Conference on
  Artificial Intelligence and Law (ICAIL 2009), Twelfth international
  conference on artificial intelligence and law (ICAIL 2009)., Barcelona,
  Spain, 8-12 June 2009}, pages 21--30. ACM.

\bibitem[\protect\citename{Nguyen and Litman}2018]{Nguyen2018ArgumentEssays}
Huy~V. Nguyen and Diane~J. Litman.
\newblock 2018.
\newblock {Argument mining for improving the automated scoring of persuasive
  essays}.
\newblock {\em 32nd AAAI Conference on Artificial Intelligence, AAAI 2018},
  pages 5892--5899.

\bibitem[\protect\citename{{OECD}}2018]{OECD2018The2030}
{OECD}.
\newblock 2018.
\newblock {The Future of Education and Skills - Education 2030}.

\bibitem[\protect\citename{{\"{O}}stling \bgroup et al.\egroup
  }2013]{Ostling2013AutomatedSwedish}
Robert {\"{O}}stling, André Smolentzov, Björn Tyrefors~Hinnerich, and Erik
  H{\"{o}}glin.
\newblock 2013.
\newblock {Automated Essay Scoring for Swedish}.
\newblock {\em Proceedings of the Eighth Workshop on Innovative Use of NLP for
  Building Educational Applications}, pages 42--47.

\bibitem[\protect\citename{Palau and Moens}2009]{Palau09}
Raquel~Mochales Palau and Marie-Francine Moens.
\newblock 2009.
\newblock Argumentation mining: The detection, classification and structure of
  arguments in text.
\newblock In {\em Proceedings of the 12th International Conference on
  Artificial Intelligence and Law}, ICAIL '09, pages 98--107, New York, NY,
  USA. ACM.

\bibitem[\protect\citename{Perelman}1971]{Perelman1971}
Chaim Perelman.
\newblock 1971.
\newblock {The New Rhetoric}.
\newblock In Yehoshua Bar-Hillel, editor, {\em Pragmatics of Natural
  Languages}, pages 145--149. Springer Netherlands, Dordrecht.

\bibitem[\protect\citename{Persing and Ng}2016]{Persing2016End-to-EndEssays}
Isaac Persing and Vincent Ng.
\newblock 2016.
\newblock {End-to-End Argumentation Mining in Student Essays}.
\newblock In {\em NAACL HLT}, pages 1384--1394.

\bibitem[\protect\citename{Pollock}1995]{pollock1995}
John~L Pollock.
\newblock 1995.
\newblock {\em {Cognitive Carpentry: A Blueprint for How to Build a Person}}.
\newblock MIT Press, Cambridge, MA, USA.

\bibitem[\protect\citename{Reed \bgroup et al.\egroup }2008a]{Reed08}
Chris Reed, Raquel Mochales~Palau, Glenn Rowe, and Marie-Francine Moens.
\newblock 2008a.
\newblock {Language Resources for Studying Argument}.
\newblock In Bente Maegaard Joseph Mariani Jan Odijk Stelios Piperidis
  Daniel~Tapias Nicoletta Calzolari (Conference Chair) Khalid~Choukri, editor,
  {\em Proceedings of the Sixth International Conference on Language Resources
  and Evaluation (LREC'08)}, Marrakech, Morocco. European Language Resources
  Association (ELRA).

\bibitem[\protect\citename{Reed \bgroup et al.\egroup
  }2008b]{Reed2008LanguageArgument}
Chris Reed, Raquel~Mochales Palau, Glenn Rowe, and Marie~Francine Moens.
\newblock 2008b.
\newblock {Language resources for studying argument}.
\newblock {\em Proceedings of the 6th International Conference on Language
  Resources and Evaluation, LREC 2008}, pages 2613--2618.

\bibitem[\protect\citename{Rietsche and
  S{\"{o}}llner}2019]{Rietsche2019InsightsSkill}
Roman Rietsche and Matthias S{\"{o}}llner.
\newblock 2019.
\newblock {Insights into Using IT-Based Peer Feedback to Practice the Students
  Providing Feedback Skill}.
\newblock {\em Proceedings of the 52nd Hawaii International Conference on
  System Sciences}.

\bibitem[\protect\citename{Rooney \bgroup et al.\egroup
  }2012]{Rooney2012ApplyingMining}
Niall Rooney, Hui Wang, and Fiona Browne.
\newblock 2012.
\newblock {Applying kernel methods to argumentation mining}.
\newblock {\em Proceedings of the 25th International Florida Artificial
  Intelligence Research Society Conference, FLAIRS-25}, pages 272--275.

\bibitem[\protect\citename{Sardianos \bgroup et al.\egroup }2015]{Sardianos15}
Christos Sardianos, Ioannis~Manousos Katakis, Georgios Petasis, and Vangelis
  Karkaletsis.
\newblock 2015.
\newblock Argument extraction from news.
\newblock In {\em Proceedings of the 2nd Workshop on Argumentation Mining},
  pages 56--66, Denver, Colorado. Association for Computational Linguistics.

\bibitem[\protect\citename{Seaman \bgroup et al.\egroup
  }2018]{Seaman2018HigherGroup}
Julia~E. Seaman, I.~E. Allen, and Jeff Seaman.
\newblock 2018.
\newblock {Higher Education Reports - Babson Survey Research Group}.
\newblock Technical report.

\bibitem[\protect\citename{Song \bgroup et al.\egroup
  }2014]{song-etal-2014-applying}
Yi~Song, Michael Heilman, Beata Beigman~Klebanov, and Paul Deane.
\newblock 2014.
\newblock Applying argumentation schemes for essay scoring.
\newblock In {\em Proceedings of the First Workshop on Argumentation Mining},
  pages 69--78, Baltimore, Maryland, June. Association for Computational
  Linguistics.

\bibitem[\protect\citename{Stab and Gurevych}2014a]{Stab2014AnnotatingEssays}
Christian Stab and Iryna Gurevych.
\newblock 2014a.
\newblock {Annotating Argument Components and Relations in Persuasive Essays}.
\newblock In {\em Proceedings of COLING 2014, the 25th International Conference
  on Computational Linguistics: Technical Papers ,}, pages 1501--1510.

\bibitem[\protect\citename{Stab and Gurevych}2014b]{Stab2014IdentifyingEssays}
Christian Stab and Iryna Gurevych.
\newblock 2014b.
\newblock {Identifying Argumentative Discourse Structures in Persuasive
  Essays}.
\newblock In {\em Conference on Empirical Methods in Natural Language
  Processing (EMNLP 2014)(Oct. 2014), Association for Computational
  Linguistics, p.(to appear)}, pages 46--56.

\bibitem[\protect\citename{Stab and Gurevych}2017a]{Stab2017ParsingEssays}
Christian Stab and Iryna Gurevych.
\newblock 2017a.
\newblock {Parsing Argumentation Structures in Persuasive Essays}.
\newblock {\em Computational Linguistics}, 43(3):619--659, 9.

\bibitem[\protect\citename{Stab and
  Gurevych}2017b]{stab-gurevych-2017-recognizing}
Christian Stab and Iryna Gurevych.
\newblock 2017b.
\newblock Recognizing insufficiently supported arguments in argumentative
  essays.
\newblock In {\em Proceedings of the 15th Conference of the {E}uropean Chapter
  of the Association for Computational Linguistics: Volume 1, Long Papers},
  pages 980--990, Valencia, Spain, April. Association for Computational
  Linguistics.

\bibitem[\protect\citename{Stenetorp \bgroup et al.\egroup
  }2012]{stenetorp-etal-2012-brat}
Pontus Stenetorp, Sampo Pyysalo, Goran Topi{\'c}, Tomoko Ohta, Sophia
  Ananiadou, and Jun{'}ichi Tsujii.
\newblock 2012.
\newblock brat: a web-based tool for {NLP}-assisted text annotation.
\newblock In {\em Proceedings of the Demonstrations at the 13th Conference of
  the {E}uropean Chapter of the Association for Computational Linguistics},
  pages 102--107, Avignon, France, April. Association for Computational
  Linguistics.

\bibitem[\protect\citename{Toulmin}1984]{Toulmin1984IntroductionReasoning}
Stephen~E. Toulmin.
\newblock 1984.
\newblock {\em {Introduction to Reasoning}}.

\bibitem[\protect\citename{Van~Eemeren and
  Grootendorst}2016]{van2016argumentation}
Frans~H Van~Eemeren and Rob Grootendorst.
\newblock 2016.
\newblock {\em {Argumentation, communication, and fallacies: A
  pragma-dialectical perspective}}.
\newblock Routledge.

\bibitem[\protect\citename{vom Brocke \bgroup et al.\egroup
  }2018]{vomBrocke2018FutureSystems}
Jan vom Brocke, Wolfgang Maa{\ss}, Peter Buxmann, Alexander Maedche, Jan~Marco
  Leimeister, and Günter Pecht.
\newblock 2018.
\newblock {Future Work and Enterprise Systems}.
\newblock {\em Business and Information Systems Engineering}, 60(4):357--366.

\bibitem[\protect\citename{Vygotsky}1980]{vygotsky1980mind}
Lev~Semenovich Vygotsky.
\newblock 1980.
\newblock {\em {Mind in society: The development of higher psychological
  processes}}.
\newblock Harvard university press.

\bibitem[\protect\citename{Wachsmuth \bgroup et al.\egroup }2014]{Wachsmuth14}
Henning Wachsmuth, Martin Trenkmann, Benno Stein, Gregor Engels, and Tsvetomira
  Palakarska.
\newblock 2014.
\newblock A review corpus for argumentation analysis.
\newblock In {\em Proceedings of the 15th International Conference on
  Computational Linguistics and Intelligent Text Processing - Volume 8404},
  CICLing 2014, pages 115--127, New York, NY, USA. Springer-Verlag New York,
  Inc.

\bibitem[\protect\citename{Walton and Macagno}2015]{Walton2015-WALACS-4}
Douglas Walton and Fabrizio Macagno.
\newblock 2015.
\newblock {A Classification System for Argumentation Schemes}.
\newblock {\em New Publisher: Ios Press}, 6(3):219--245.

\bibitem[\protect\citename{Walton \bgroup et al.\egroup
  }2008]{Walton2008ArgumentationSchemes}
Douglas Walton, Christopher Reed, and Fabrizio Macagno.
\newblock 2008.
\newblock {\em {Argumentation Schemes}}.
\newblock Cambridge University Press, Cambridge.

\bibitem[\protect\citename{Walton}1996]{walton1996argumentation}
D~N Walton.
\newblock 1996.
\newblock {\em {Argumentation Schemes for Presumptive Reasoning}}.
\newblock Argumentation Schemes for Presumptive Reasoning. L. Erlbaum
  Associates.

\bibitem[\protect\citename{Wambsganss and
  Rietsche}2020]{Wambsganss2019TowardsTool}
Thiemo Wambsganss and Roman Rietsche.
\newblock 2020.
\newblock {Towards designing an adaptive argumentation learning tool}.
\newblock In {\em 40th International Conference on Information Systems, ICIS
  2019}, pages 1--9, Munich, Germany.

\bibitem[\protect\citename{Wambsganss \bgroup et al.\egroup
  }2020a]{Wambsganss2020UnlockingApproach}
Thiemo Wambsganss, Nikolaos Molyndris, and Matthias S{\"{o}}llner.
\newblock 2020a.
\newblock {Unlocking Transfer Learning in Argumentation Mining : A
  Domain-Independent Modelling Approach}.
\newblock In {\em 15th International Conference on Wirtschaftsinformatik,},
  number~Ml, Potsdam, Germany.

\bibitem[\protect\citename{Wambsganss \bgroup et al.\egroup
  }2020b]{Wambsganss2020ALSkills}
Thiemo Wambsganss, Christina Niklaus, Matthias Cetto, Matthias S{\"{o}}llner,
  Jan~Marco Leimeister, and Siegfried Handschuh.
\newblock 2020b.
\newblock {AL : An Adaptive Learning Support System for Argumentation Skills}.
\newblock In {\em ACM CHI Conference on Human Factors in Computing Systems},
  pages 1--14.

\bibitem[\protect\citename{Weinberger and
  Fischer}2006]{Weinberger2006ALearning}
Armin Weinberger and Frank Fischer.
\newblock 2006.
\newblock {A framework to analyze argumentative knowledge construction in
  computer-supported collaborative learning}.
\newblock {\em Computers and Education}, 46(1):71--95.

\bibitem[\protect\citename{Yannakoudakis \bgroup et al.\egroup
  }2011]{Yannakoudakis2011ATexts}
Helen Yannakoudakis, Ted Briscoe, and Ben Medlock.
\newblock 2011.
\newblock {A new dataset and method for automatically grading ESOL texts}.
\newblock {\em ACL-HLT 2011 - Proceedings of the 49th Annual Meeting of the
  Association for Computational Linguistics: Human Language Technologies},
  1:180--189.

\end{thebibliography}

\end{document}